\pdfoutput=1

\documentclass[11pt]{article}

\usepackage[final]{acl}

\usepackage{times}
\usepackage{latexsym}

\usepackage[T1]{fontenc}

\usepackage[utf8]{inputenc}

\usepackage{microtype}

\usepackage{inconsolata}

\usepackage{graphicx}

\usepackage{microtype}
\usepackage{hyperref}
\usepackage{url}
\usepackage{booktabs}

\usepackage{lineno}

\definecolor{darkblue}{rgb}{0, 0, 0.5}
\hypersetup{colorlinks=true, citecolor=darkblue, linkcolor=darkblue, urlcolor=darkblue}

\usepackage{float}

\usepackage{tabularx}
\newcolumntype{C}{>{\centering\arraybackslash}X} 

\usepackage{xcolor}
\usepackage{booktabs}
\usepackage{colortbl}

\usepackage{tcolorbox}
\tcbuselibrary{breakable}

\usepackage{mdframed}
\usepackage{multirow}
\usepackage{expex}
\usepackage{amsmath}
\usepackage[whole]{bxcjkjatype}
\usepackage{subcaption}
\usepackage{pifont}
\usepackage{wrapfig}
\usepackage{etoolbox}
\usepackage{titlesec}
\usepackage{authblk}
\usepackage{tikz}

\usepackage{fontawesome5}
\usepackage{hyperref}

\title{Pragmatic Theories Enhance Understanding of Implied Meanings in LLMs}

\author[1,2,a]{\textbf{Takuma Sato}}
\author[2,3]{\textbf{Seiya Kawano}}
\author[1,2,4]{\textbf{Koichiro Yoshino}}

\affil[1]{Nara Institute of Science and Technology, Nara, Japan}
\affil[2]{Guardian Robot Project, RIKEN}
\affil[3]{Kyoto Institute of Technology}
\affil[4]{Institute of Science Tokyo}

\affil[a]{\texttt{sato.takuma.sq6@naist.ac.jp}}

\begin{document}
\maketitle
\begin{abstract}
\renewcommand{\thefootnote}{\fnsymbol{footnote}}
\footnotetext{
This is the preprint version of the paper accepted to the main conference of IJCNLP-AACL 2025. 
Our code is available at \faGithub ~\url{https://github.com/takuma1229/pragmatic-theories-enhance}.}
The ability to accurately interpret implied meanings plays a crucial role in human communication and language use, and language models are also expected to possess this capability.
This study demonstrates that providing language models with pragmatic theories as prompts is an effective approach for tasks to understand implied meanings.
Specifically, we propose an approach in which an overview of pragmatic theories, such as Gricean pragmatics and Relevance Theory, is presented as a prompt to the language model, guiding it through a step-by-step reasoning process to derive a final interpretation.
Experimental results showed that, compared to the baseline, which prompts intermediate reasoning without presenting pragmatic theories (0-shot Chain-of-Thought), our methods enabled language models to achieve up to 9.6\% higher scores on pragmatic reasoning tasks.
Furthermore, we show that even without explaining the details of pragmatic theories, merely mentioning their names in the prompt leads to a certain performance improvement (around 1-3\%) in larger models.
\end{abstract}

\section{Introduction}

Language often contains implicit meanings, unstated intentions, and context-dependent interpretations, collectively referred to as \textbf{implied meanings}. 
The ability to correctly understand and interpret implied meanings plays a crucial role in human communication \cite{grice1989-studies-in-the-way-of-words, Levinson1983-pragmatics, Carston2002-thoughts-and-utterances}.
In order to correctly interpret the implied meanings, people use background knowledge of communication, such as dialogue context, common sense, and cultural background.
Using these, they achieve effective communication in their daily lives.
The ability to understand implied meanings is important not only for humans but also for AI systems or robots based on them.
For example, when a human's intention is ambiguous, inferring their intentions and taking actions proactively while considering the situation and common sense is necessary \cite{tanaka2024-do-as-i-demand}.
The ability to interpret implied meaning, as well as understanding the nuances and intentions of language, is essential for AI and robots to be needed in human society.

In recent years, following the success of Large Language Models (LLMs), research has been conducted on employing them to develop systems that have abilities to interpret language and situations~\cite{ahn2022-do-as-i-can, ma2025-pragmatics-in-the-era-of-large-language}.
Numerous benchmark tasks and datasets have been proposed to assess these capabilities \cite{jeretic2020-are-natural-language-inference, zheng2021-grice-a-grammar-based, takayama2021-direct, hu2023-fine-grained-comparison, li2023-diplomat, sravanthi2024-pub-a-pragmatics, yerukola2024-is-the-pope-catholic-yes, shisen2024-do-large-language-models-chinese-sitcom}, facilitating comparisons between current language models and human performance, as well as analyses of the challenges these models face.

Two major approaches have been explored to enhance the understanding of implied meanings in LLMs: post-training involving parameter updates~\cite{wu2024-rethinking-pragmatics, sravanthi2024-pub-a-pragmatics}, and in-context learning (ICL), which draws out specific capabilities of LLMs through carefully designed prompts~\cite{yerukola2024-is-the-pope-catholic-yes, ruis2023-the-goldilocks-of-pragmatic-understanding, kim2023-is-pope-catholic-applying}. 
ICL is particularly important, as it enables the full utilization of the model's abilities without additional training costs, and has been addressed with extensive approaches regarding its generalization capabilities. 
In the context of understanding implied meanings, there is a growing need to explore not only instance-specific methods but also more instance-agnostic methods.

In this study, we propose a method that enhances model performance on tasks to understand implied meanings in a zero-shot setting without relying on prompts specific to particular problem formats or providing top-down hints from correct answers.
Implied meanings are handled in the field of linguistics known as \textbf{pragmatics}, where various theories have been proposed regarding their properties and the mechanisms of their interpretation.
The proposed method inserts summaries of theoretical frameworks from linguistic pragmatics into the model's prompt and instructs the model to generate intermediate reasoning processes by following those theories.
It is known that existing LLMs have a variety of knowledge in their parameters; however, making them recall and use this knowledge appropriately for the task is still challenging. 
By giving the rough outline of pragmatic theories as a bootstrap, we expected that these models could recall the related knowledge to be used and manipulate them for solving such tasks to understand the implied meanings.

Experimental results demonstrated that the proposed method consistently improved model performance on tasks to understand implied meanings without providing instance-dependent information, achieving an accuracy improvement of up to 0.096 in the experimental tasks.
We tested the proposed method in both commercial-closed and open models and found that the proposed method contribute to a wide range of models. 
Additionally, slight score improvements were observed in many models even when the prompt did not include an overview of pragmatic theories but only mentioned their names while encouraging reasoning by following them.
These investigations not only have engineering utility and can be expected to apply to upstream tasks such as dialogue, but also lead to clarifying the nature of the task of pragmatic understanding itself and what kind of thought processes are effective for executing such tasks.
Our contributions can be summarized as follows:
\begin{itemize}
    \vspace{-5pt}
    \item We proposed \textbf{a simple method that incorporates summaries of pragmatic theories (namely Gricean and Relevance Theories) into prompts} and showed that this improves LLM performance on implied meaning understanding tasks without preparing task-dependent prompts.
    \vspace{-5pt}
    \item We showed that even without explaining the theories in detail, \textbf{simply referencing theory names} in zero-shot Chain-of-Thought prompting leads to performance improvements over baseline methods.
    \vspace{-5pt}
    \item Through detailed quantitative and qualitative analyses, we demonstrated that our method is particularly effective for tasks involving utterances that flout Grice's maxims and for interpreting irony.
\end{itemize}

\section{Related Works}
    \subsection{Pragmatic Reasoning by Language Models}
    Previous research has demonstrated that methods such as post-training, Chain-of-Thought (CoT) reasoning, and few-shot learning effectively enhance the pragmatic inference capabilities of language models.
    As post-training approaches using pragmatic reasoning datasets, policy optimization methods such as Direct Preference Optimization (DPO) are more effective than supervised fine-tuning \cite{wu2024-rethinking-pragmatics}.
    It has also been confirmed that instruction-tuned models achieve higher scores in pragmatic reasoning tasks compared to their corresponding base models \cite{ruis2023-the-goldilocks-of-pragmatic-understanding, sravanthi2024-pub-a-pragmatics}.
    These methods involve updating the model's parameters.

    For in-context learning methods that do not involve updating the model's parameters, previous research has shown that CoT prompting, where guidances for correct interpretation are included in the prompt \cite{yerukola2024-is-the-pope-catholic-yes}, and few-shot prompting, where the prompts provide examples of problems and correct answers similar to the target task \cite{ruis2023-the-goldilocks-of-pragmatic-understanding}, are effective approaches.
    As a method that combines these approaches, experiments have reported that providing reasoning steps based on Gricean theory \cite{grice1989-studies-in-the-way-of-words} within few-shot examples can serve as effective guidance for correct interpretation \cite{kim2023-is-pope-catholic-applying}.

    However, there has been insufficient research on methods for enhancing the pragmatic reasoning capabilities of language models without ad hoc interventions, such as modifying model parameters or providing top-down hints specific to the task.
    For large-scale pretrained LLMs, conducting post-training with parameter updates (e.g., supervised fine-tuning or preference optimization) is not only costly and labor-intensive but also carries the risk of ``catastrophic forgetting,'' where the performance on previously learned tasks deteriorates after fine-tuning on a specific task \cite{kirkpatrick2017-overcoming-catastrophic-forgetting, li2024-revisiting-catastrophic-forgetting}.
    Additionally, existing in-context learning methods should be developed in light of the fact that pragmatic inference is often not an end goal in itself but rather a necessary component ability for higher-level tasks or objectives.
    Given this, the approaches in prior studies \cite{yerukola2024-is-the-pope-catholic-yes, ruis2023-the-goldilocks-of-pragmatic-understanding, kim2023-is-pope-catholic-applying}, which rely on top-down and ad hoc applications of few-shot learning or CoT prompting tailored to specific pragmatic reasoning problems, may not be sufficient.
    We tackle their remaining challenge to maintain generalized prompts for interpreting implied meanings using pragmatic theories.

    \subsection{Pragmatic Theories in Linguistics}
    We propose a method based on two well-established pragmatic theories from the fields of linguistics and philosophy of language: Gricean pragmatics \cite{grice1989-studies-in-the-way-of-words} and Relevance Theory \cite{sperber1996-relevance}.

    \paragraph*{Gricean Theory}
    Grice proposed a pragmatic theory in which he argued that the correct interpretation of what a speaker means in an utterance is achieved by assuming that participants in a conversation adhere to, or at least appear to adhere to, the \textit{Cooperative Principle}.
    This principle serves as the foundation for the reasoning made by listeners.

    \vspace{-3pt}
    \begin{quote}
        \textbf{\textit{Cooperative Principle} \cite{grice1989-studies-in-the-way-of-words}}\\
            Make your conversational contribution such as is required, at the stage at which it occurs, by the accepted purpose or direction of the talk exchange in which you are engaged (p.26).
    \end{quote}
    \vspace{-3pt}
    
    He also argued that adherence to the cooperative principle requires the observance of the four subordinate maxims, namely \textit{Maxim of Quantity, Quality, Relation, and Manner} (see Appendix \S\ref{appendix:the_gricean_maxim}).

    Grice explains the understanding and interpretation of implied meanings based on the idea that our conversations generally adhere to these maxims.
    When they do not, the listener infers meaning by recognizing that they are violating one or more maxims while assuming that the speaker still follows the \textit{cooperative principle}.
    Grice's theory has had a significant impact not only in linguistics and philosophy of language but also in Natural Language Processing (NLP), where it has been widely applied~\cite{krause2024-gricean-maxims}.

    Though Grice’s work founded modern pragmatics, later research exposed its limits and inspired refinements such as Relevance Theory, which offers a more cognitively plausible account of utterance interpretation \cite{levinson2000-presumptive-meanings, horn1984-towards-a-new-taxonomy, wearing2015-relevance-theory-pragmatics}.

    \paragraph*{Relevance Theory}
    A key feature of Relevance Theory, proposed by Sperber and Wilson, is its cognitive approach to pragmatic meaning \cite{sperber1996-relevance}.
    Relevance Theory defines \textit{relevance} in terms of two factors: the \textit{cognitive effect}, which refers to the degree to which an utterance influences the listener's thoughts, and the \textit{processing effort} required to understand or interpret the utterance.
    All else being equal, an utterance is considered more relevant if it produces greater \textit{cognitive effects} and requires less \textit{processing effort}.
    Relevance Theory asserts that, based on this notion of \textit{relevance}, we expect the \textit{presumption of relevance} (see Appendix \S \ref{appendix:presumption_of_relevance}) in communication and that pragmatic meaning is interpreted accordingly.

    Our study proposes a method that incorporates summaries of Gricean pragmatics and Relevance Theory into the model's prompt, guiding the model to generate reasoning processes aligned with these theories.

    \subsection{Zero-shot prompt templates}
    \label{related_works:zero-shot_prompt_templates}
    It is known that specifying certain thinking methods as prompts can generically improve model performance without giving language models problem-answer pairs~\cite{brown-2020-language-models-are-few-shot} or providing explicit hints~\cite{wei2022-chain-of-thought-prompting-elicits}.
    \citet{kojima2022-large-language-models-are-zero-shot} showed that simply inputting the text ``Let's think step by step.'' to a model can significantly improve performance on benchmarks for various tasks, and this method is called \textit{zero-shot Chain-of-Thought}. Such methods are not only convenient for practical use of language models, but also noteworthy for exploring the foundations and mechanisms that realize their capabilities, so various studies have been conducted subsequently~\cite{wang2023-plan-and-solve, wang2023-self-consistency-improves-chain-of-thought, schulhoff2025-the-prompt-report}.

    In such contexts, several prompt templates have been discovered that aim to maximize LLM capabilities in specific tasks or domains. 
    For example, \citet{he2024-exploring-human-like-translation} shows that prompt templates that make models extract keywords and topics from source sentences before performing final translation improve LLM performance on translation tasks. 
    Additionally, \citet{sonish2024-an-empirical-evaluation} demonstrated that zero-shot prompt templates based on medical domain knowledge are effective in clinical information extraction tasks. 
    Methods of \textit{jailbreak attacks}~\cite{yi2024-jailbreak-attacks-defenses-large}, which are intended to elicit harmful or inappropriate outputs from models by devising prompts, might also be positioned within this context.

\section{Experiments}
    \subsection{Dataset and Task}
    We use PRAGMEGA \cite{floyd2022-pragmega-materials, hu2023-fine-grained-comparison} as a pragmatic reasoning dataset and task. This dataset covers seven broad pragmatic phenomena observed in English conversations. 
    Each instance in the dataset requires correctly answering questions designed primarily to test understanding an utterance's implied meanings, considering the utterance itself and its conversational context.

    Each problem in the dataset is classified into one of the seven pragmatic phenomena the dataset addresses.
    These seven pragmatic phenomena are \textit{Deceits}, \textit{Indirect speech}, \textit{Irony}, \textit{Maxims}, \textit{Metaphor}, \textit{Humor}, and \textit{Coherence}.
    Since the present study specifically focuses on the task of correctly interpreting non-literal meanings in utterances, we conducted experiments targeting five of these pragmatic phenomena, excluding \textit{Humor} and  \textit{Coherence}, which do not directly address such meanings.
    The following descriptions are provided in \cite{hu2023-fine-grained-comparison} for each phenomenon\footnote{Examples of problems corresponding to each phenomenon are shown in Table \ref{table:pragmega_phenomenon_examples} in the appendix.}:
    \begin{itemize}
        \item \textbf{\textit{Deceits}}: \textit{Polite deceits} used for social or personal relationships. In the questions, respondents must employ \textit{Theory of Mind} and other reasoning skills to determine why the speaker used a particular utterance correctly.
        \vspace{-5pt}
        \item \textbf{\textit{Indirect Speech}}: Utterances with performative meanings, such as prompting others to take action. The questions test whether respondents understand what the speaker tries to convey in stories involving indirect requests.
        \vspace{-15pt}
        \item \textbf{\textit{Irony}}: Utterances that communicate the opposite of their literal meaning. The questions require interpreting what the characters intend to convey through the presented irony.
        \vspace{-5pt}
        \item \textbf{\textit{Maxims}}: Utterances that violate one of Grice's four maxims. Respondents determine why the characters made such utterances.
        \vspace{-5pt}
        \item \textbf{\textit{Metaphor}}: Utterances that depict comparisons between entities in a non-literal manner. The questions present metaphors within a story and ask respondents to interpret what the speaker is trying to convey.
    \end{itemize}

    The total number of problems is 520, comprising 100 instances each for \textit{Deceits}, \textit{Indirect Speech}, and \textit{Metaphor}, 125 instances for \textit{Irony}, and 95 instances for \textit{Maxims}.
    Our experiments do not explicitly indicate the phenomenon to which each problem belongs within the prompt.  
    Instead, a common prompt format is used across all phenomena to present the problem and response structure.

    \subsection{Methods}
    We conducted comparative experiments on the following prompting methods (two baselines and two proposed methods).
    \color{gray}Baseline-1/2\color{black} ~denote baseline methods, and 
    \color[HTML]{b5af3d}Proposed-1/2\color{black} ~denote proposed methods.
    The baseline and proposed methods are instance-agnostic, meaning they do not provide models with instance-dependent information or top-down hints.
    Within these instance-agnostic methods, we hypothesize that our proposed methods, including summaries of pragmatic theory in prompts, would achieve higher performance on pragmatic reasoning tasks than the baseline methods. 
    This is because, by indicating these theories, we expect the model to know which area of knowledge or reasoning contained in the model should be called.
    Note that all experiments adopt a zero-shot learning setup, meaning no task-solving examples are provided in the prompts.

    \begin{figure*}[htbp]
        \centering
        \includegraphics[width=0.9\linewidth]{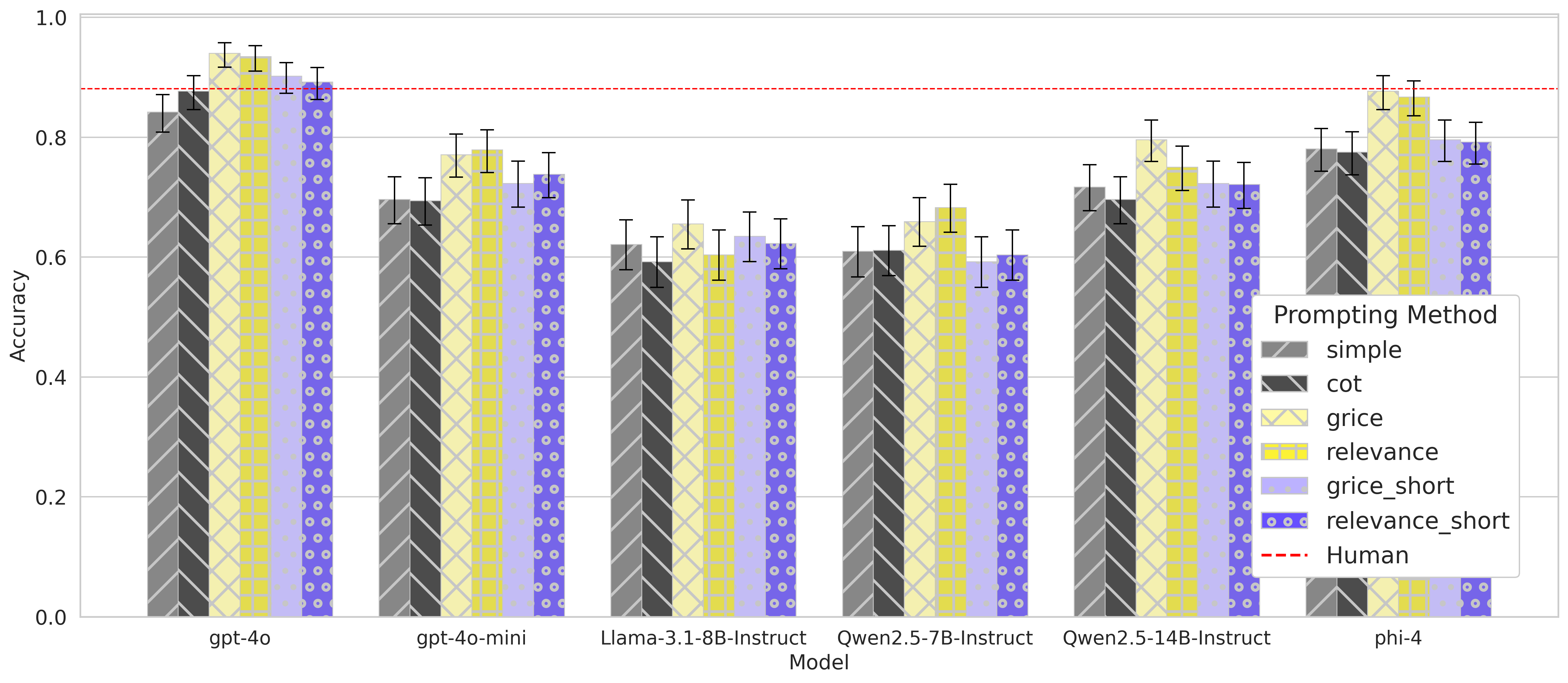}
        \caption{Accuracies on pragmatic inference task of PRAGMEGA. In most models, the proposed methods outperformed the baseline methods.
        The human scores indicate scores presented in the original paper by \cite{hu2023-fine-grained-comparison}.  
        Error bars represent 95\% confidence intervals calculated using Wilson's method~\cite{Wilson1927-probable-inference-the-law-of}.
        Even with a short prompt for the pragmatic theory, larger models showed improvements from the proposed methods; however, the extent of improvement was smaller compared to when the theory was explained in detail.
        }
        \label{fig:accuracy}
    \end{figure*}

    \begin{tcolorbox}[colback=lightgray!20, colframe=gray, boxrule=1pt, breakable]
    \paragraph*{\color{gray}Baseline-1\color{black} : Simple}
    A setting in which the prompts explicitly instruct the models to output \textbf{only} the final selected answer.

    \paragraph*{\color{gray}Baseline-2\color{black} : Chain-of-Thought (\texttt{cot})}
    A setting in which the model is prompted with ``\texttt{Firstly, think step-by-step and write down your process of thinking},'' explicitly instructing it to output its reasoning process leading to the final answer, followed by the selected answer.
    This method is often referred to as \textit{zero-shot Chain-of-Thought} \cite{kojima2022-large-language-models-are-zero-shot}\footnote{Note that this method differs from the ``Chain-of-Thought'' approach presented in \cite{yerukola2024-is-the-pope-catholic-yes} and \cite{kim2023-is-pope-catholic-applying}, as it does not provide instance-specific hints for correct interpretation.}.

    \paragraph*{\color[HTML]{b5af3d}Proposed-1\color{black} : Gricean Prompting (\texttt{grice})}
    A setting in which the prompt provides the models with a brief overview of Gricean theory \cite{grice1989-studies-in-the-way-of-words}, explicitly instructing them to output a reasoning process aligned with this overview, followed by the final selected answer.

    \paragraph*{\color[HTML]{b5af3d}Proposed-2\color{black} : Relevance Theory Prompting (\texttt{relevance})}
    A setting in which the prompt provides the models with a brief overview of Relevance Theory \cite{sperber1996-relevance, Carston2002-thoughts-and-utterances}, explicitly instructing them to describe a reasoning process aligned with this overview before outputting the final selected answer.
    
    \end{tcolorbox}

    \subsection{Models}
    We conducted experiments using various LLMs implementing decoder-based Transformer architecture \cite{Vaswani2017-attention-is-all-you-need, liu2018-generating-wikipedia}.
    We experimented with publicly available models (open models), whose source code and pre-trained parameters are released, and proprietary models (closed models), whose parameters are not publicly available.
    As Open models, we selected the LLaMa3 series \cite{grattafiori2024-the-llama3-herd-of-models} and the Qwen2.5 series \cite{qwen2025-qwen2.5-technical-report}.
    As Closed models, we selected GPT-4o and GPT-4o mini \cite{openai2024-gpt4omini}.
    Considering the length of our prompts, we chose models with sufficiently large context lengths (16k tokens or more) for the experiments.
    Appendix \S\ref{appendix:hyperparameters} shows the hyperparameters used in the experiments.
    About computation details, see \S \ref{appendix:computation_details} in the Appendix.

\section{Results}
    \subsection{Experimental Results}
    \label{subsection:experimental_results}

    The experimental results shown in Figure \ref{fig:accuracy} confirm that the proposed methods perform better than the baseline methods on the pragmatic reasoning task\footnote{
    The exact scores are presented in \S \ref{appendix:experimental_results}, \ref{appendix:exact_result_by_phenomona} in the Appendix.
    }.
    Notably, with GPT-4o, the proposed methods enabled the model to achieve performance surpassing that of humans.
    Even in the case of phi-4, where there was a performance gap between the baseline methods and human scores, applying our methods allowed the model to reach human-level accuracy.

    Using \textbf{Gricean prompting} (\texttt{grice}) resulted in higher scores than both baseline methods across all models tested.
    This method had the most significant effect on phi-4, where accuracy improved by 0.096 compared to the higher-scoring baseline method (\texttt{simple}).
    Even for Llama-3.1-8B-Instruct, where the most minor improvement was observed, accuracy increased by 0.032.

    For \textbf{Relevance prompting} (\texttt{relevance}), no performance improvement was observed for Llama-3.1-8B-Instruct compared to the baseline, but all other models showed performance gains.
    When comparing \texttt{grice} and \texttt{relevance}, \texttt{grice} achieved higher scores in most models.
    While some models recorded higher scores with \texttt{relevance}, the difference from \texttt{grice} in those cases was minimal.

    For \textbf{Short prompting} (\texttt{grice short, relevance short}), performance improvements over the baseline were observed in all models except Qwen2.5-7B-Instruct.
    Even in models where scores improved over the baseline, the magnitude of improvement was generally smaller than that achieved by the proposed methods.

    \subsection{Analysis}

    \subsubsection{General Overview}
    \label{subsubsection:general_overview}
    When comparing \texttt{grice} and \texttt{relevance}, \texttt{grice} achieved higher scores in most models.
    While some models recorded higher scores with \texttt{relevance}, the difference from \texttt{grice} in those cases was minimal.
    However, this difference does not necessarily indicate that Gricean Theory is more valid than Relevance Theory.
    Instead, we speculate that the difference is due to Gricean Theory appearing more frequently in the training corpus of the models.

    There was no clear trend indicating that longer input or output lengths consistently led to higher accuracy.
    An analysis of the relationships between input/output length and accuracy revealed little correlation between these factors.
    The Pearson correlation coefficients between accuracy and input length and accuracy and output length were 0.181 and 0.211, respectively.
    Additionally, the coefficient of determination ($R^2$) from simple regression analysis was 0.032 and 0.044, respectively (see Appendix \S\ref{appendix:correlation_analysis} for detailed analysis results).

    \subsubsection{Analyses by Pragmatic Phenomena}
    \begin{figure*}[htbp]
        \centering
        \includegraphics[width=0.9\linewidth]{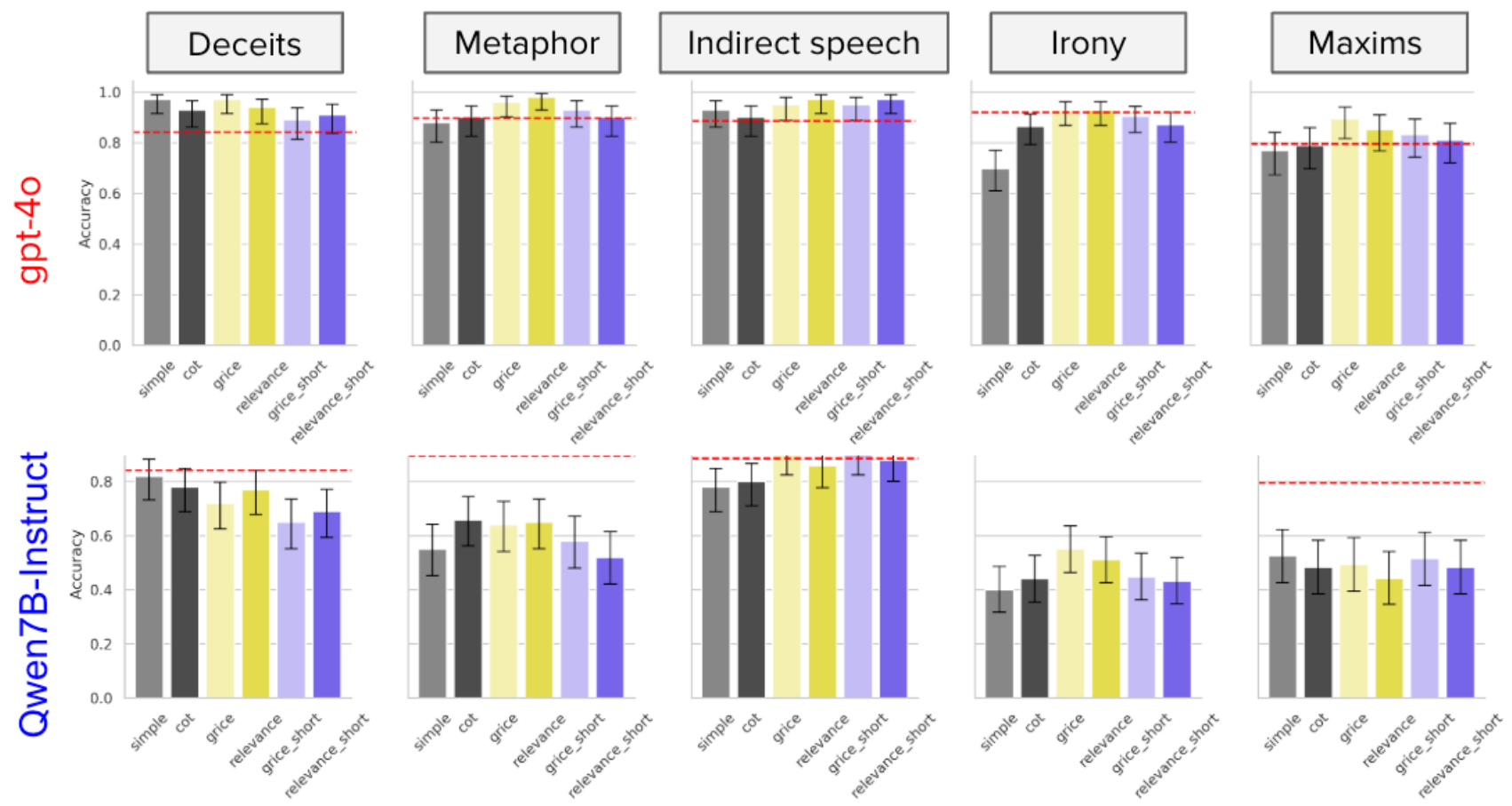}
        \caption{
        Accuracy of the model for each pragmatic phenomenon included in PRAGMEGA \cite{hu2023-fine-grained-comparison} when using different methods.  
        Due to space constraints, we present the results for GPT-4o and Qwen2.5-7B-Instruct (for detailed results, including other models, see Appendix \S\ref{appendix:exact_result_by_phenomona}).
        The human score is based on \cite{hu2023-fine-grained-comparison}.  
        }
        \label{fig:by_phenomena}
    \end{figure*}

    A summary of the analysis results for each of the five pragmatic phenomena included in PRAGMEGA, comparing different prompting methods using GPT-4o and Qwen2.5-7B-Instruct, is shown in Figure \ref{fig:by_phenomena}.
    \footnote{For exact scores, see \S \ref{appendix:exact_result_by_phenomona} in the appendix.
    }
    The results showed the most notable score improvement from the proposed methods in \textit{Irony}.
    In the case of GPT-4o, baseline methods resulted in performance lower than that of humans; however, by applying the proposed methods, the model achieved human-level accuracy.
    For this phenomenon, regardless of whether the prompts provide the models with an overview of the theory, using Gricean Theory generally resulted in higher scores than Relevance Theory.
    In \textit{Indirect Speech}, although the margin was smaller than in \textit{Irony}, consistent improvements over the baseline methods were also observed.
    However, we found no clear superiority between Gricean Theory and Relevance Theory in this phenomenon.
    For \textit{Maxims}, using Gricean Prompting, which includes an overview of Grice's maxims, did not always lead to improved scores over the baseline.
    However, for models with 14B or more parameters, the proposed methods generally resulted in higher accuracy.
    In most cases, models performed better when using Gricean Theory compared to Relevance Theory, and this trend was consistent even in the short prompts experiments.
    For \textit{Metaphor}, models with 14B or more parameters tended to improve scores when using the proposed methods.
    In GPT-4o, GPT-4o-mini, and Qwen2.5-14B-Instruct, performance slightly improved with at least one of the proposed methods, even when using short prompts, though no performance improvement was observed with short prompts in phi-4.
    For \textit{Deceits}, the effectiveness of the proposed methods varied depending on the model, and we observed no clear or consistent trend in relation to models' parameter size.

    \subsubsection{Error Analysis}
    \label{subsubsection:error_analysis}

    \begin{figure}[tbp]
        \centering
        \includegraphics[width=1\linewidth]{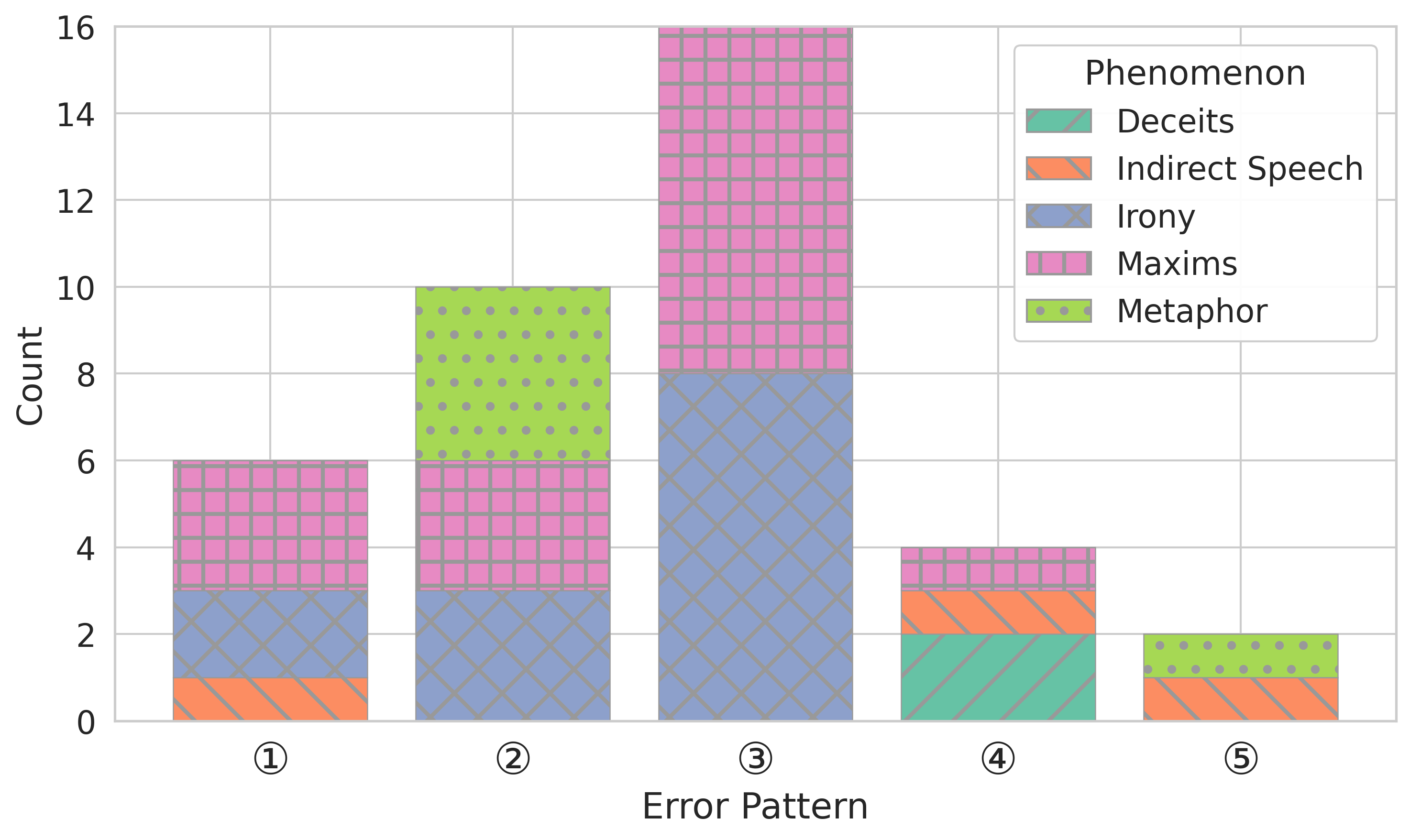}
        \caption{
        The number of instances for each error pattern by GPT-4o, as described in the main text.  
        A cumulative bar chart represents these counts, including the distribution of each phenomenon within each pattern.
        }
        \label{fig:error_pattern_bar_chart}
    \end{figure}

    We conducted error analysis using the results from GPT-4o, categorized errors into the following five patterns, and counted the number of instances corresponding to each phenomenon (Figure \ref{fig:error_pattern_bar_chart}). 
    For \textcircled{\scriptsize 1} and \textcircled{\scriptsize 3}, actual examples of errors are presented in Table \ref{table:error_examples_in_body}.
    Examples of the other error patterns are provided in Appendix Table \ref{table:error_examples_in_appendix}.

    \vspace{-5pt}
    \paragraph{\textcircled{\scriptsize 1} Cases where the proposed methods were clearly effective}  
    This category includes cases where \texttt{simple} and \texttt{cot} produced incorrect answers, but all other methods resulted in correct answers.  
    Many problems classified under \textit{Maxims} fell into this category.  
    The first example in Table \ref{table:error_examples_in_body} corresponds to this pattern.  
    We hypothesize that including pragmatic theories in the prompt made the model more sensitive to the distinction between \textit{what is said} and \textit{what is implied}.

    \vspace{-5pt}
    \paragraph{\textcircled{\scriptsize 2} Cases where \textit{Short Prompting} was insufficient}  
    This category includes cases where \texttt{grice} and \texttt{relevance} resulted in correct answers, but all other methods produced incorrect answers.  
    A relatively large proportion of problems in this pattern fell under \textit{Metaphor}.  
    This suggests that understanding metaphors may not benefit from a superficial grasp of pragmatic theories alone, but a more detailed comprehension of these theories could enable their application to interpretation in such cases.  

    \vspace{-5pt}
    \paragraph{\textcircled{\scriptsize 3} Cases where all methods failed}  
    In this category, problems classified under \textit{Maxims} and \textit{Irony} appeared in equal numbers.  
    Since Figure \ref{fig:by_phenomena} also indicates that these two phenomena were the most challenging for the models, it is natural that they appear prominently in this error pattern.  
    Indeed, the second example in Table \ref{table:error_examples_in_body} seems difficult even for humans.  
    Furthermore, determining that option 4 is the correct answer might require additional contextual information.  
    If such problems can be considered to lack a ``definitive correct answer,'' then the performance of cutting-edge models like GPT-4o may already be saturated within our experimental setting.

    \paragraph{\textcircled{\scriptsize 4} Cases where only \textit{Gricean Prompting} was effective}  
    This category includes cases where only \texttt{grice} and \texttt{grice\_short} resulted in correct answers, while all other methods failed.  
    The most frequent phenomenon observed in this pattern was \textit{Deceits}.  
    It could be valuable to explore how different results might emerge if theories that focus more on social aspects, such as Politeness Theory \cite{Brown1987-politeness}, were incorporated into the prompts.

    \paragraph{\textcircled{\scriptsize 5} Cases where only \textit{Relevance Prompting} was effective}  
    This category includes cases where only \texttt{relevance} and \texttt{relevance\_short} resulted in correct answers, while all other methods failed.  
    This pattern was relatively rare, with only two instances equally distributed between the \textit{Metaphor} and \textit{Indirect Speech} phenomena.  

    \begin{table*}[tbp]
        \centering
        \fontsize{9.1pt}{9.1pt}\selectfont
        \caption{
        Actual question examples for some error patterns.
        \color{blue}\textbf{Bold}\color{black} ~indicates the correct options.
        Due to space constraints, examples other than \textcircled{\scriptsize 1} and \textcircled{\scriptsize 3} are provided in Appendix \S\ref{appendix:examples_of_each_error_pattern}.
        }
        
        \begin{tabularx}{\textwidth}{p{1cm}|X|X}
        \hline
        \rowcolor{gray!10} 
        \textbf{Pattern} & \textbf{Questions} & \textbf{Options} \\ \hline \hline
        
        \textcircled{\scriptsize 1} &
        Samantha is talking with her dad about her fiance. Samantha notes: ``John is an innocent person.'' Her dad replies: ``Undoubtedly, as innocent as a saint.'' Why has Samantha's dad responded like this? &1. Samantha's dad is impressed with John's innocence. \newline
        \textbf{\color{blue}2. Samantha's dad thinks that Samantha has an incorrect view of her fiance.\color{black}} \newline
        3. Samantha's dad thinks that Samantha's fiance is a saint. \newline
        4. Samantha's dad thinks that John is too religious. \\ \hline

        \textcircled{\scriptsize 3} & John is a teacher at an elementary school. When talking with the principal about a new student, who did poorly on her entrance examination, John said, ``This one is really sharp.'' What did John want to convey? & 1. The entrance exam is unfair. \newline 2. The pencils need to be sharpened. \newline 3. The student is smart. \newline \textbf{\color{blue}4. The student is not very clever.\color{black}} \\ \hline

        \end{tabularx}
        \label{table:error_examples_in_body}
    \end{table*}
    
\section{Additional Experiments: Examination of Confounding Factors}
\label{section:additional_experiment}

\subsection{Motivation and Experimental Settings}
\label{subsection:additional_experiment:motivation_and_settings}
In the previous section, we demonstrated that providing an explanation of pragmatic theory can improve the performance of various LLMs on the PRAGMEGA dataset, and furthermore, that this improvement is not merely a result of input or output length effects.  
We also hypothesized that our prompt might have effectively activated the LLMs' latent knowledge of pragmatic theory.   
However, the improvement observed with the proposed method may stem from confounding factors rather than the intended pragmatic theories.
Specifically, the following concerns remain\footnote{These possibilities were pointed out in comments from anonymous reviewers during the ARR 2025 July Cycle.}.
\vspace{-5pt}
\begin{enumerate}
    \item Even when using a theory that is authoritative in appearance but largely unrelated to the task—rather than a pragmatic theory—the performance might improve to a similar or even greater extent than with the proposed method.
    \vspace{-16pt}
    \item Even when using an entirely fictitious ``pragmatic theory'' that is nonsensical but appears formally plausible, the performance might improve to a similar or even greater extent than with the proposed method.
    \vspace{-5pt}
    \item Even without using a pragmatic theory, simply using a general prompt template other than 0-shot CoT might yield performance improvements comparable to or greater than those achieved by the proposed method.
\end{enumerate}

To test these concerns, we ran additional experiments with different prompts for each scenario.  
Foremost, to investigate the first concern, we conducted an experiment using the following methods, which incorporate a theory unrelated to the task.
\begin{tcolorbox}[colback=lightgray!20, colframe=gray, boxrule=1pt, breakable]
    \paragraph*{X-bar Theory Prompting (\texttt{xbar})}
    In this setting, the prompt instructs the model to ``think in line with X-bar theory~\cite{Chomsky1957-syntactic-structure, chomsky1968-remarks-on-nominalization}.''  
    \vspace{5pt}

    \paragraph*{Computational Complexity Prompting (\texttt{complexity})}
    In this setting, the prompt instructs the model to ``think in line with the theory of computational complexity.''
    \vspace{5pt}

    \paragraph*{Graph Theory Prompting (\texttt{graph})}
    In this setting, the prompt instructs the model to ``think in line with the graph theory.''
\end{tcolorbox}

\begin{figure*}[tbp]
        \centering
        \includegraphics[width=1\linewidth]{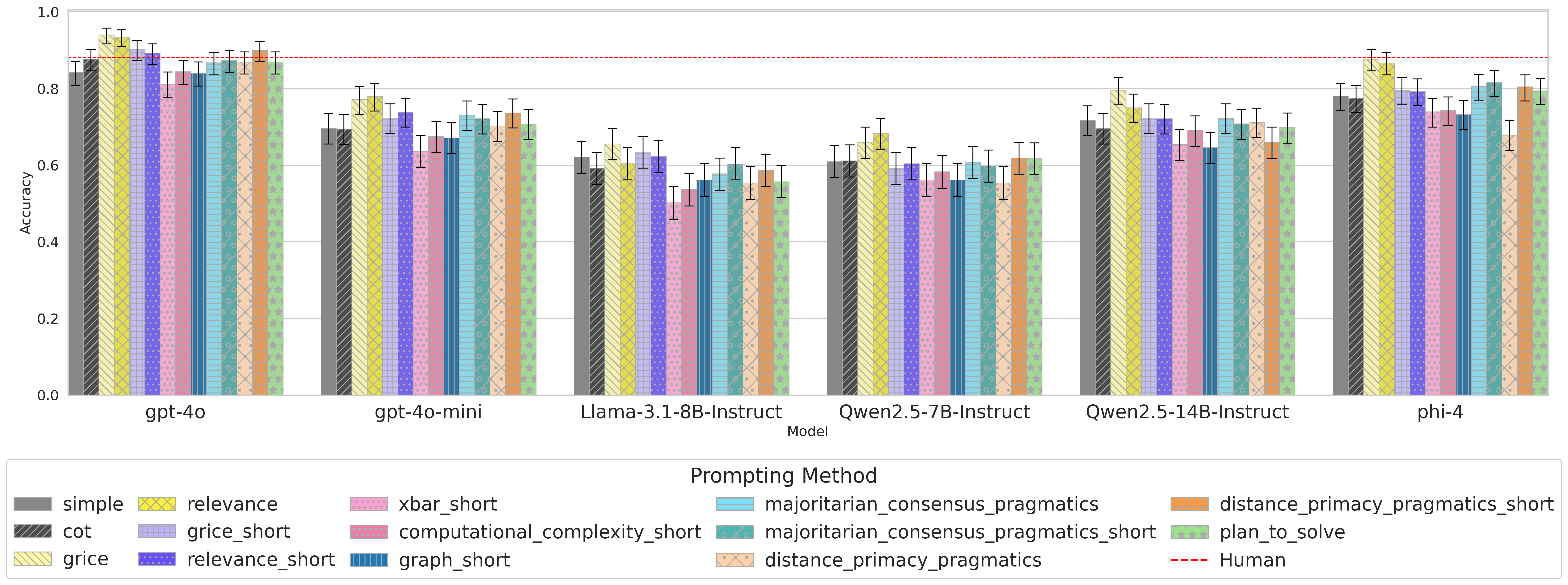}
        \caption{Results of the additional experiments}
        \label{fig:additional_experiments_result}
\end{figure*}

To test the second concern, we used fictitious pragmatic theories in the following experiments.
\begin{tcolorbox}[colback=lightgray!20, colframe=gray, boxrule=1pt, breakable]
    \paragraph*{Majoritarian Pragmatics Prompting (\texttt{majoritarian})}
    In this setting, the model receives a brief description of a made-up pragmatic theory, Majoritarian Consensus Pragmatics, and is told to reason according to it. 
    This theory claims that an expression's meaning is simply what the majority would interpret it as.
    \vspace{5pt}

    \paragraph*{Majoritarian Pragmatics Short Prompting (\texttt{majoritarian\_short})}
    In this setting, the prompt instructs the model to ``think in line with Majoritarian Consensus Pragmatics,'' but the description of the theory itself is omitted from the prompt.
    \vspace{5pt}

    \paragraph*{Distance Pragmatics Prompting (\texttt{distance})}
    Here, the model is told to reason and answer using a brief description of the fictional theory named Distance Primacy Pragmatics.
    This theory claims that the primary factor in utterance interpretation is the physical distance (spatial parameter) between speaker and listener, while linguistic content and social context are merely secondary.
    \vspace{5pt}

    \paragraph*{Distance Pragmatics Short Prompting (\texttt{distance\_short})}
    In this setting, the prompt instructs the model to ``think in line with Distance Primacy Pragmatics,'' but the description of the theory itself is omitted from the prompt.
    \vspace{5pt}
\end{tcolorbox}
Additionally, to examine the third concern, we conducted experiments using the following method.  
\begin{tcolorbox}[colback=lightgray!20, colframe=gray, boxrule=1pt, breakable]
    \paragraph*{Plan-to-Solve Prompting (\texttt{plan})}
    This setting employs the Plan-to-Solve Prompting method proposed by \citet{wang2023-plan-and-solve}.  
    In the prompt, the model is instructed as follows:  
    ``Let’s first understand the problem and devise a plan to solve it. Then, let’s carry out the plan and solve the problem step by step.''
\end{tcolorbox}

The specific prompts used for each method are presented in Tables~\ref{table:prompting-methods-in-additional-experiments-1}, \ref{table:prompting-methods-in-additional-experiments-2}, and \ref{table:prompting-methods-in-additional-experiments-3} in Appendix~\ref{appendix:prompting_method}.  
The models and detailed experimental settings are identical to those described in the previous section.

\subsection{Results and Analysis}
\label{subsection:additional_experiments:results_and_analysis}

The results of the additional experiments are shown in Figure~\ref{fig:additional_experiments_result}.  
Detailed results are presented in Table~\ref{table:additional_exps_exact_results} in the appendix.  
Among all combinations of models and additional prompting methods, none achieved a higher accuracy than the proposed methods, \texttt{grice} and \texttt{relevance}.  
In other words, no experimental evidence was found to support the concerns above.  

However, among the additional methods, some achieved higher accuracy than \texttt{grice\_short} and \texttt{relevance\_short}.  
In particular, \texttt{distance\_short} frequently exhibited such cases.  
This may suggest that even if a theory is entirely fictitious in name, the model can sometimes find useful meaning or interpretation in the prompt and solve the task, as long as the prompt is not absurd or incoherent.

Providing prompts based on well-known but task-irrelevant theories---whose content the model is likely to have encountered during pretraining---tended instead to decrease the model’s performance on our task.  
Furthermore, using a general prompt template, such as \texttt{plan}, sometimes resulted in either decreased or improved performance compared to the baseline methods.

\section{Conclusion}
In this paper, we proposed an instance-agnostic method for pragmatic reasoning tasks by incorporating an overview of linguistic pragmatic theories into the model's prompt.
Experimental results demonstrated that the proposed methods improve performance on pragmatic reasoning tasks without providing instance-dependent information to the model in a top-down manner.  

As a direction for future work, it is desirable to develop pragmatic reasoning tasks that require consideration of richer and more extended contexts, followed by experiments using such tasks.
A limitation in PRAGMEGA and other previous studies is that the instances they handle often lack sufficient complexity compared to real-world pragmatic phenomena, or the contextual information relevant to interpretation is insufficiently rich.
To address these issues, creating datasets that incorporate more complex and contextually rich pragmatic phenomena, along with tasks utilizing these datasets, is necessary for a more precise analysis of model capabilities and limitations.
The dataset used in \cite{shisen2024-do-large-language-models-chinese-sitcom}, which is based on a Chinese sitcom, represents a promising approach to addressing these challenges.  
Focusing on this direction and developing more comprehensive datasets and benchmarks 
will contribute significantly to the further advancement of this research field.


Finally, exploring the approach of applying discussions from specific domains as prompt templates, as we have done in other domains and tasks, would contribute to optimizing model performance strategies and exploring the nature of the tasks themselves in those use cases.

\section*{Limitations}
One limitation of our research is that we have not been able to verify whether the proposed method is effective when applied to more upstream tasks or broader language domains using language models. 
For example, if we could demonstrate the generalized effectiveness of the proposed method in the performance of other tasks where pragmatic abilities such as question answering, dialogue, and instruction following could be useful, the value of this research would be further enhanced. 

Additionally, a limitation of this research is that we have not sufficiently verified ``why'' the proposed method was effective in the experiments. 
More advanced verification is needed to elucidate why this method improves performance and why it does not improve all phenomena. 
For example, we stated in Sec. \ref{subsubsection:general_overview} that the hypothesis is that the differences in score trends across theories might be due to the frequency of data included in the training; however, more detailed analysis is required to make definitive statements about this.

It is also important to verify whether we can obtain similar results with datasets in languages other than English.
Since the incorporation and degree of pragmatic meaning in language use vary across languages \cite{Baumgarten2022-contrastive-pragmatics}, assessing the consistency of our method's effectiveness across diverse languages would be valuable.

\section*{Acknowledgments}
A part of this work was supported by JST PRESTO grant number JPMJPR24TC.
Also, our work was partly supported by the Toyota Foundation, grant number D24-ST-0023.
Furthermore, the additional experiments described in Section~\ref{section:additional_experiment} were conducted with the valuable advice of anonymous reviewers in the ACL Rolling Review July 2025 Cycle.  
We would like to express our sincere gratitude for their contributions.

\section*{Ethics Statements}
\paragraph{AI Assistants}
In this study, AI assistants, including ChatGPT, Copilot, and DeepL, were used in accordance with the ACL Policy on AI Writing Assistance. 
We primarily used them to assist with coding and writing, but all code and text outputs were manually reviewed. 
The authors take full responsibility for all of them.


\bibliography{custom}

\appendix

\section{The Gricean Maxim}
\label{appendix:the_gricean_maxim}
Grice argued that to adhere to the Cooperative Principle, the following four subordinate maxims must be observed.
\begin{quote}
    \textbf{\textit{The Gricean Maxims} \cite{grice1989-studies-in-the-way-of-words}}
    \begin{enumerate}
        \item \textit{Maxim of Quantity}: Provide an amount of information that is neither too little nor too much.
        \item \textit{Maxim of Quality}: Do not say what you believe to be false or for which you lack sufficient evidence.
        \item \textit{Maxim of Relation}: Be relevant in your utterance.
        \item \textit{Maxim of Manner}: Avoid ambiguity and be clear, concise, and orderly in your expression.
    \end{enumerate}
\end{quote}

\section{Presumption of Relevance}
\label{appendix:presumption_of_relevance}
\begin{quote}
    \textbf{\textit{Presumption of Relevance} \cite{sperber1996-relevance}}
    \begin{enumerate}
        \item An ostensive stimulus is relevant enough to make it worth the addressee's effort to process it.
        \item An ostensive stimulus is the most relevant one, given the communicator's abilities and priorities.
    \end{enumerate}
\end{quote}

\section{Examples of PRAGMEGA dataset}
\label{appendix:pragmega_example}
Table \ref{table:pragmega_phenomenon_examples} shows examples of problems and choices for each phenomenon in the PRAGMEGA dataset.

    \begin{table*}[htbp]
        \centering
        \fontsize{9.1pt}{9.1pt}\selectfont
        \caption{Examples of problems from each pragmatic phenomenon that PRAGMEGA includes. \color{blue}\textbf{Bold}\color{black} ~indicates the correct answers. In the actual dataset, the order of choices is randomly shuffled. We adapt the table from examples shown in \cite{hu2023-fine-grained-comparison}.}
        \vspace{5pt}
        \begin{tabular}{m{2cm}|m{4.5cm}|m{6.2cm}}
        \hline
        \rowcolor{gray!10} 
        \textbf{Phenomenon} & \textbf{Questions} & \textbf{Options} \\ \hline
        
        Deceits &
        Henry is sitting at his desk and watching TV, and reluctantly switches off the TV with the remote control and picks up a textbook. Shortly after, his mother comes in the room and asks, ``What have you been doing up here?'' Henry responds: ``Reading.'' Why has Henry responded in such a way? &
        \color{blue}\textbf{1. He does not want to get into trouble for not studying.}\color{black} \newline
        2. He has been reading for some time. \newline
        3. He does not want to offend his mom by not reading the books that she gave him. \newline
        4. He wants his mom to believe that he has been watching TV. \\ \hline

        Metaphor &
        Andrew and Bob were discussing the investment company where Andrew works. Bob said: ``The investors are squirrels collecting nuts.'' What does Bob mean? &
        \color{blue}\textbf{1. They buy stocks hoping for future profit.}\color{black} \newline
        2. Squirrels were hired to work in the company. \newline
        3. The investors dress and eat well. \newline
        4. Bob is allergic to nuts. \newline
        5. The investors enjoy picking nuts as squirrels do. \\ \hline
        
        Indirect Speech &
        Nate is about to leave the house. His wife points at a full bag of garbage and asks: ``Are you going out?'' What might she be trying to convey? &
        \color{blue}\textbf{1. She wants Nate to take the garbage out.}\color{black}\newline
        2. She wants to know Nate's plans. \newline
        3. She wants Nate to bring his friends over. \newline
        4. She wants Nate to spend more time with the family. \\ \hline
        
        Irony &
        It is a holiday. Stefan and Kim are sitting in the backseat of the car. They are fighting all the time. Their father says: ``Oh, it is so pleasant here.'' What did the father want to convey? &
        \color{blue}\textbf{1. He does not want to listen to his kids' arguments.}\color{black}  \newline
        2. He enjoys listening to his kids fighting. \newline
        3. AC gives them some needed cool. \newline
        4. He remembers about his wife's birthday. \\ \hline
        
        Maxims &
        Leslie and Jane are chatting at a coffee shop. Leslie asks, ``Who was that man that I saw you with last night?'' Jane responds, ``The latte is unbelievable here.'' Why has Jane responded like this? &
        \color{blue}\textbf{1. She does not want to discuss the topic that Leslie has raised.}\color{black} \newline
        2. She thinks that it is the best latte in the town. \newline
        3. The man who Leslie saw makes unbelievable lattes. \newline
        4. A coffee break is not a good time to discuss men. \\ \hline

        \end{tabular}
        \label{table:pragmega_phenomenon_examples}
    \end{table*}

\section{Each Prompting method}
\label{appendix:prompting_method}
Table \ref{table:prompting-methods} shows each prompt used as a method in the experiments of this study.
     \begin{table*}[htbp]
        \centering
        \caption{Prompts used in the methods compared in this study.}
        \vspace{5pt}
        \small
        \renewcommand{\arraystretch}{1.5} 
        \setlength{\tabcolsep}{10pt} 
        \begin{tabularx}{\textwidth}{p{2cm}X}
            \rowcolor{gray!10} 
            \toprule
            \textbf{Method} & \textbf{Prompt} \\
            \midrule
            \raggedright\textbf{Simple} (\color{gray}Baseline-1\color{black}) & \scriptsize\texttt{Write ONLY the option number of your final answer and its contents in the format like: [Answer] 2) hogehoge is hogehoge. Any additional output beyond this penalized.} \\ 
            \midrule
            \raggedright\textbf{Chain-of-Thought} (\color{gray}Baseline-2\color{black}) & \scriptsize\texttt{Firstly, think step-by-step and write down your process of thinking. After that, select your final answer. Your final answer should be in the format like: [Answer] 2) hogehoge is hogehoge.} \\
            \midrule
            \raggedright\textbf{Gricean Prompting} (\color[HTML]{b5af3d}Proposed-1\color{black}) & \parbox[t]{\linewidth}{\scriptsize\texttt{Let's think in line with the Gricean theory.\\
            In Grice's framework, hearers arrive at the implied meanings (or ``implicatures'') through an inferential process guided by the Cooperative Principle and its associated conversational maxims (Quantity, Quality, Relation, and Manner). 
            Specifically: \\ 
            Cooperative Principle: The assumption that speakers and hearers are cooperating with one another to communicate effectively.\\ \\
            Conversational Maxims: \\
            Quantity: Be as informative as required (but not overly so).\\
            Quality: Do not say what you believe to be false or lack evidence for.\\
            Relation (Relevance): Be relevant.\\
            Manner: Be clear, avoid ambiguity and obscurity.\\ \\
            When a hearer detects a potential mismatch between what is said (literally) and one of the maxims, they hypothesize a conversational implicature—that the speaker must mean something more or different than the literal meaning. The hearer then uses context, background knowledge, and reasoning about the speaker's intent and adherence to the maxims to infer the intended meaning.\\
            Write down your thinking process in line with Gricean theory and ultimately decide on the final answer.\\
            Your final answer should be in the format like: [Answer] 2) hogehoge is hogehoge.}} \\
            \midrule
            \raggedright\textbf{Relevance Theory Prompting} (\color[HTML]{b5af3d}Proposed-2\color{black}) & \parbox[t]{\linewidth}{\scriptsize\texttt{Let's think in line with the Relevance theory.\\
            According to Relevance Theory, the interpretation of utterance implicatures proceeds through the following processes, where the balance between **cognitive effects** and **processing effort** plays a crucial role.**\\
            ---\\
            1. **Starting Point of Utterance Interpretation**:  
               - The linguistic meaning (logical form) of an utterance is merely a ``clue'' to the interpretation intended by the speaker.  
               - The listener must infer the speaker's intended meaning behind the utterance using this linguistic clue as a basis.\\
               ... \\
            6. **Interaction Between Explicit Meaning and Implicature**:  
               - Explicit meaning (the overt content of the utterance) and implicature (implied content) influence each other during processing.  
               - This interaction forms the overall interpretation of the utterance. \\ \\
            In Relevance Theory, **``optimal relevance''** is achieved when an utterance delivers **``cognitive effects worth the processing effort''** to the listener. Thus, the balance between cognitive effects and processing effort is consistently emphasized in utterance interpretation. By seeking interpretations that maximize effects with minimal effort, listeners achieve efficient understanding of utterances.\\
            Write down your thinking process in line with Relevance theory and ultimately decide on the final answer.\\
            Your final answer should be in the format like: [Answer] 2) hogehoge is hogehoge.}} \\
            \midrule
            \raggedright\textbf{Short Gricean Prompting} (\color{blue}Short-1\color{black}) & \parbox[t]{\linewidth}{\scriptsize\texttt{Let's think in line with the Gricean theory.\\ Write down your thinking process in line with Gricean theory and ultimately decide on the final answer.\\\\Your final answer should be in the format like:\\\relax [Answer] \\2) hogehoge is hogehoge.}} \\
            \midrule
            \raggedright\textbf{Short Relevance Prompting} (\color{blue}Short-2\color{black}) & \parbox[t]{\linewidth}{\scriptsize\texttt{Let's think in line with the Relevance theory.\\Write down your thinking process in line with Relevance theory and ultimately decide on the final answer.\\\\Your final answer should be in the format like:\\\relax [Answer] \\2) hogehoge is hogehoge.}} \\
            \bottomrule
        \end{tabularx}
        \label{table:prompting-methods}
    \end{table*}

    \begin{table*}[tbp]
        \centering
        \caption{Prompts used in the additional experiments to examine the first concern.}
        \small
        \renewcommand{\arraystretch}{1.5} 
        \setlength{\tabcolsep}{10pt} 
        \begin{tabularx}{\textwidth}{p{2cm}X}
            \rowcolor{gray!10} 
            \toprule
            \textbf{Method} & \textbf{Prompt} \\
            \midrule
            \raggedright\texttt{xbar} & \begin{minipage}[t]{\linewidth}
            \scriptsize\ttfamily
            Let’s think in line with the X-bar theory. Write down your thinking process in line with X-bar theory and ultimately decide on the final answer.\\\\
            Your final answer should be in the format like: \\\relax [Answer]\\ 2) hogehoge is hogehoge.
            \end{minipage} \\
            \midrule
            \texttt{complexity} & \begin{minipage}[t]{\linewidth}
            \scriptsize\ttfamily
            Let’s think in line with the computational complexity theory.
            Write down your thinking process in line with computational complexity theory and ultimately decide on the final answer. \\\\
            Your final answer should be in the format like:\\\relax
            [Answer] \\
            2) hogehoge is hogehoge.
            \end{minipage} \\
            \midrule
            \texttt{graph} & \begin{minipage}[t]{\linewidth}
            \scriptsize\ttfamily
            Let’s think in line with the graph theory. Write down your thinking process in line with the graph theory and ultimately decide on the final answer.\\\\
            Your final answer should be in the format like:\\\relax
            [Answer] \\
            2) hogehoge is hogehoge.
            \end{minipage} \\ 
            \bottomrule
        \end{tabularx}
        \label{table:prompting-methods-in-additional-experiments-1}
    \end{table*}
    
    \begin{table*}[tbp]
        \centering
        \caption{Prompts used in the additional experiments to examine the second concern.}
        \small
        \renewcommand{\arraystretch}{1.5} 
        \setlength{\tabcolsep}{10pt} 
        \begin{tabularx}{\textwidth}{p{2cm}X}
            \rowcolor{gray!10} 
            \toprule
            \textbf{Method} & \textbf{Prompt} \\
            \midrule
            \texttt{majoritarian} & \begin{minipage}[t]{\linewidth}
            \scriptsize\ttfamily
            Let’s think in line with Majoritarian Concensus Pragmatics.\\\\
            Majoritarian Consensus Pragmatics (MCP) posits that the meaning of an expression is always what the greatest number of people take it to mean. Individual intention or dictionary-based definition is secondary; what truly defines pragmatic meaning is the distribution of social agreement. Meaning is not “correct” in an objective sense but rather “dominant” by virtue of majority interpretation.\\\\
            MCP's key concepts are summaried below:\\
            1. Majoritarian Meaning
            The interpretation of an utterance is determined by the intuition of the majority within a speech community. Minority readings may exist but are pragmatically considered “non-meaningful.”\\
            2. Consensus Index
            Every expression carries a “consensus index,” a hypothetical measure of how many people favor a given interpretation at a certain moment. For example, if 90\% interpret “yabai” as “cool,” that becomes the pragmatically valid meaning.\\
            3. Floating Meaning
            Since consensus shifts over time, meanings are inherently unstable. This accounts for semantic shifts, slang evolution, and the inversion of meaning (e.g., “yabai” changing from negative to positive).\\\\
            Write down your thinking process in line with Majoritarian Concensus Pragmatics and ultimately decide on the final answer.\\\\
            Your final answer should be in the format like:\\\relax
            [Answer] \\
            2) hogehoge is hogehoge.
            \end{minipage} \\
            \midrule
            \texttt{majoritarian short} & \begin{minipage}[t]{\linewidth}
            \scriptsize\ttfamily
            Let’s think in line with Majoritarian Concensus Pragmatics.
            Write down your thinking process in line with Majoritarian Concensus Pragmatics and ultimately decide on the final answer.\\\\
            Your final answer should be in the format like:\\\relax
            [Answer] \\
            2) hogehoge is hogehoge.
            \end{minipage} \\
            \midrule
            \texttt{distance} & \begin{minipage}[t]{\linewidth}
            \scriptsize\ttfamily
            Let’s think in line with Distance-Primacy Pragmatic Theory.\\\\
            The Distance-Primacy Pragmatic Theory (DPPT) posits that the primary determinant of utterance interpretation is the “spatial parameter” between speaker and hearer, while linguistic content and social context are treated as secondary. Utterances are processed according to the Distance-Meaning Mapping Model, which assigns distinct pragmatic functions to specific ranges of physical distance. Key concepts of this theory are summaried below:\\
            1. Principle of Proximal Compulsion (PPC)
            Any utterance produced within a radius of 50 cm is automatically interpreted as a command, regardless of the speaker’s actual intent.\\
            2. Mid-Distance Propositional Zone (MDPZ)
            Within the range of 1--3 meters, utterances are processed as proposals or invitations.\\
            3.Phenomenon of Distant Monologization (PDM)
            Beyond 3 meters, all utterances are treated as pragmatic monologues, releasing the hearer from any obligation to respond.\\\\
            Write down your thinking process in line with Distance-Primacy Pragmatic Theory and ultimately decide on the final answer.\\
            Your final answer should be in the format like:\\\relax
            [Answer] \\
            2) hogehoge is hogehoge.
            \end{minipage} \\
            \midrule
            \texttt{distance short} & \begin{minipage}[t]{\linewidth}
            \scriptsize\ttfamily
            Let’s think in line with Distance-Primacy Pragmatic Theory.
            Write down your thinking process in line with Distance-Primacy Pragmatic Theory and ultimately decide on the final answer.
            \\\\Your final answer should be in the format like:\\\relax [Answer] \\ 2) hogehoge is hogehoge.
            \end{minipage} \\
            \bottomrule
        \end{tabularx}
        \label{table:prompting-methods-in-additional-experiments-2}
    \end{table*}
    
    \begin{table*}[tbp]
        \centering
        \caption{Prompts used in the additional experiments to examine the third concern.}
        \small
        \renewcommand{\arraystretch}{1.5} 
        \setlength{\tabcolsep}{10pt} 
        \begin{tabularx}{\textwidth}{p{2cm}X}
            \rowcolor{gray!10} 
            \toprule
            \textbf{Method} & \textbf{Prompt} \\
            \midrule
            \raggedright\texttt{plan} & \begin{minipage}[t]{\linewidth}
            \scriptsize\ttfamily
            Let’s first understand the problem and devise a plan to solve the problem. Then, let's carry out the plan and solve the problem step by step.\\\\
            Your final answer should be in the format like:\\\relax
            [Answer] \\
            2) hogehoge is hogehoge.
            \end{minipage} \\
            \bottomrule
        \end{tabularx}
        \label{table:prompting-methods-in-additional-experiments-3}
    \end{table*}

\section{Hyperparameters}
\label{appendix:hyperparameters}
We set \texttt{temperature=0.8, max\_new\_tokens=1500, repetition\_penalty=1.2, do\_sample=True}.

\section{Computation Details}
\label{appendix:computation_details}
The parameter counts for each open model we used are as follows: 
\begin{itemize}
    \item Llama-3.1-8B-Instruct: 8B
    \vspace{-5pt}
    \item Qwen2.5-7B-instruct: 7B
    \vspace{-5pt}
    \item Qwen2.5-14B-instruct: 14B
    \vspace{-5pt}
    \item phi-4: 14B
\end{itemize}
We used NVIDIA RTX™ A6000, and our experiments took around 200 GPU hours.
All models were quantized to 8-bit before performing inference using vLLM \cite{kwon2023-efficient-memory-management}. 
The experiments were conducted on March 2, 2025.

\section{Exact Experimental Results}
\label{appendix:experimental_results}
In our main experiments, the exact accuracy achieved by each model is shown in \ref{tab:exact_results_overall}.
\begin{table*}[htbp]
        \caption{Main experimental results. The highest Accuracy among the four methods is indicated in \textbf{bold}.}
        \vspace{5pt}
        \centering
        \small
        \begin{tabular}{|c|cc|cc|cc|}
            \toprule
            \textbf{Model} & \multicolumn{2}{c|}{\textbf{Baseline}} & \multicolumn{2}{c|}{\textbf{Proposed}} & \multicolumn{2}{c|}{\textbf{Short}}\\
            \cmidrule(lr){2-3} \cmidrule(lr){4-7} 
            &  \textbf{\texttt{simple}} & \textbf{\texttt{cot}} & \textbf{\texttt{grice}} & \textbf{\texttt{relevance}} & \textbf{\texttt{grice}} & \textbf{\texttt{relevance}} \\
            \midrule
            \midrule
            gpt-4o  & 0.842 & 0.877 & \textbf{0.940} & 0.935 & 0.902 & 0.892 \\
            gpt-4o-mini  & 0.696 & 0.694 & 0.771 & \textbf{0.779} & 0.723 & 0.738 \\
            Llama-3.1-8B-Instruct  & 0.621 & 0.592 & \textbf{0.656} & 0.604 & 0.635 & 0.623 \\
            Qwen2.5-7B-Instruct  & 0.610 & 0.612 & 0.660 & \textbf{0.683} & 0.592 & 0.604 \\
            Qwen2.5-14B-Instruct  & 0.717 & 0.696 & \textbf{0.796} & 0.750 & 0.723 & 0.721 \\
            phi-4  & 0.781 & 0.775 & \textbf{0.877} & 0.867 & 0.796 & 0.792\\
            \bottomrule
        \end{tabular}
        \label{tab:exact_results_overall}
\end{table*}

In addition, the detailed results for each method used in the additional experiments are presented in Table~\ref{table:additional_exps_exact_results}.

\begin{table*}[tbp]
    \centering
    \caption{Exact experiment results for the additional experiment}
    \label{table:additional_exps_exact_results}
    \begin{tabular}{l||r|r|r|r|r|r}
    \toprule
    \rowcolor{gray!20}
     & gpt-4o & gpt-4o-mini & Llama8B & Qwen7B & Qwen14B & phi-4 \\
    \midrule
    \midrule
    \texttt{simple} & 0.842 & 0.696 & 0.621 & 0.610 & 0.717 & 0.781 \\
    \texttt{cot} & 0.877 & 0.694 & 0.592 & 0.612 & 0.696 & 0.775 \\
    \texttt{grice} & 0.940 & 0.771 & 0.656 & 0.660 & 0.796 & 0.877 \\
    \texttt{relevance} & 0.935 & 0.779 & 0.604 & 0.683 & 0.750 & 0.867 \\
    \texttt{grice\_short} & 0.902 & 0.723 & 0.635 & 0.592 & 0.723 & 0.796 \\
    \texttt{relevance\_short} & 0.892 & 0.738 & 0.623 & 0.604 & 0.721 & 0.792 \\
    \texttt{xbar} & 0.812 & 0.637 & 0.502 & 0.562 & 0.654 & 0.739 \\
    \texttt{complexity} & 0.844 & 0.675 & 0.537 & 0.583 & 0.691 & 0.743 \\
    \texttt{graph} & 0.840 & 0.671 & 0.561 & 0.562 & 0.647 & 0.733 \\
    \texttt{majoritarian} & 0.867 & 0.731 & 0.578 & 0.608 & 0.722 & 0.807 \\
    \texttt{majoritarian\_short} & 0.873 & 0.721 & 0.603 & 0.598 & 0.708 & 0.815 \\
    \texttt{distance} & 0.869 & 0.702 & 0.554 & 0.554 & 0.712 & 0.678 \\
    \texttt{distance\_short} & 0.900 & 0.737 & 0.587 & 0.619 & 0.660 & 0.804 \\
    \texttt{plan} & 0.869 & 0.708 & 0.557 & 0.617 & 0.698 & 0.794 \\
    \hline
    \end{tabular}
\end{table*}

\section{Correlation Analyses between Accuracy and Input/output Length}
\label{appendix:correlation_analysis}
\begin{figure*}[htbp]
        \centering
        \includegraphics[width=0.9\linewidth]{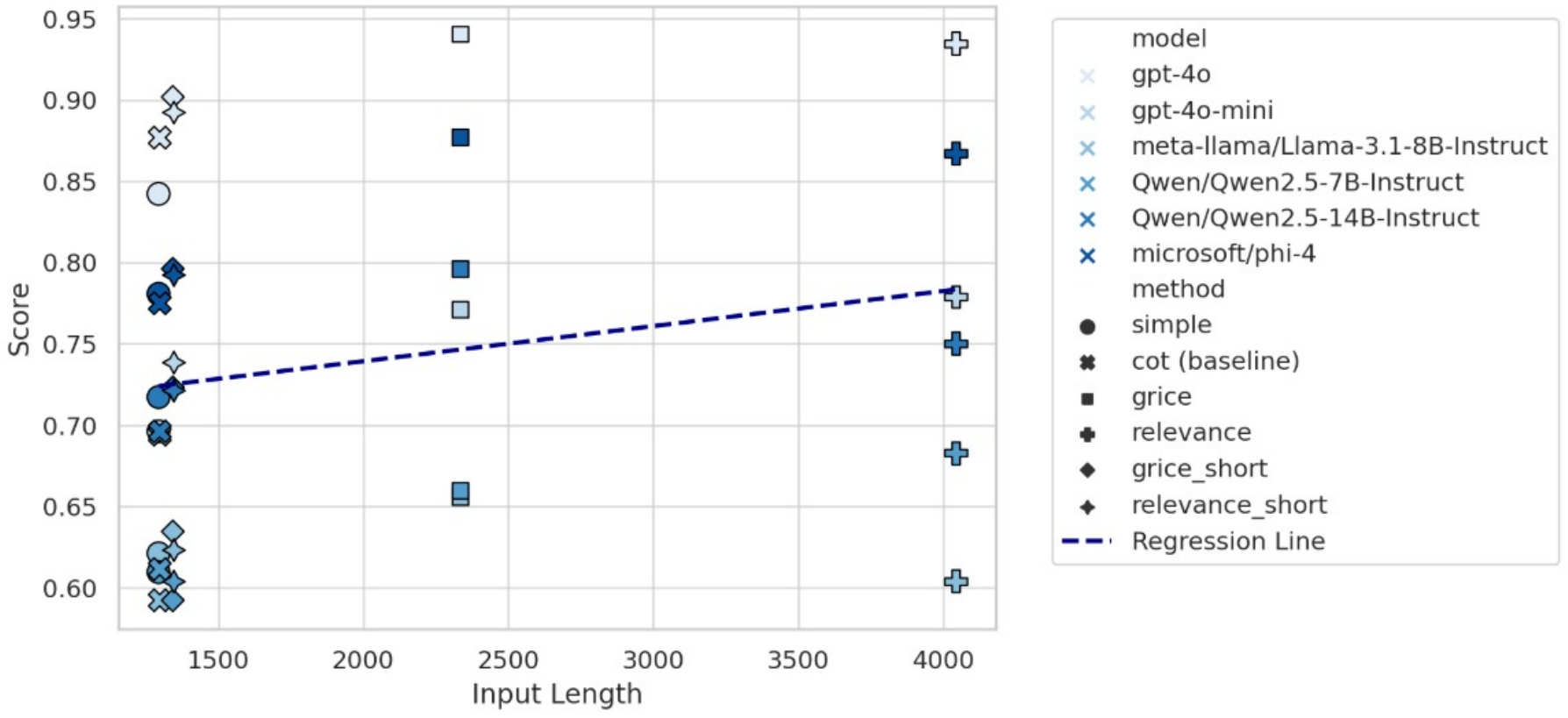}
        \caption{Correlation analysis between input length and accuracy.}
        \label{fig:correlation_analysis_input}
    \end{figure*}

\begin{figure*}[htbp]
        \centering
        \includegraphics[width=0.9\linewidth]{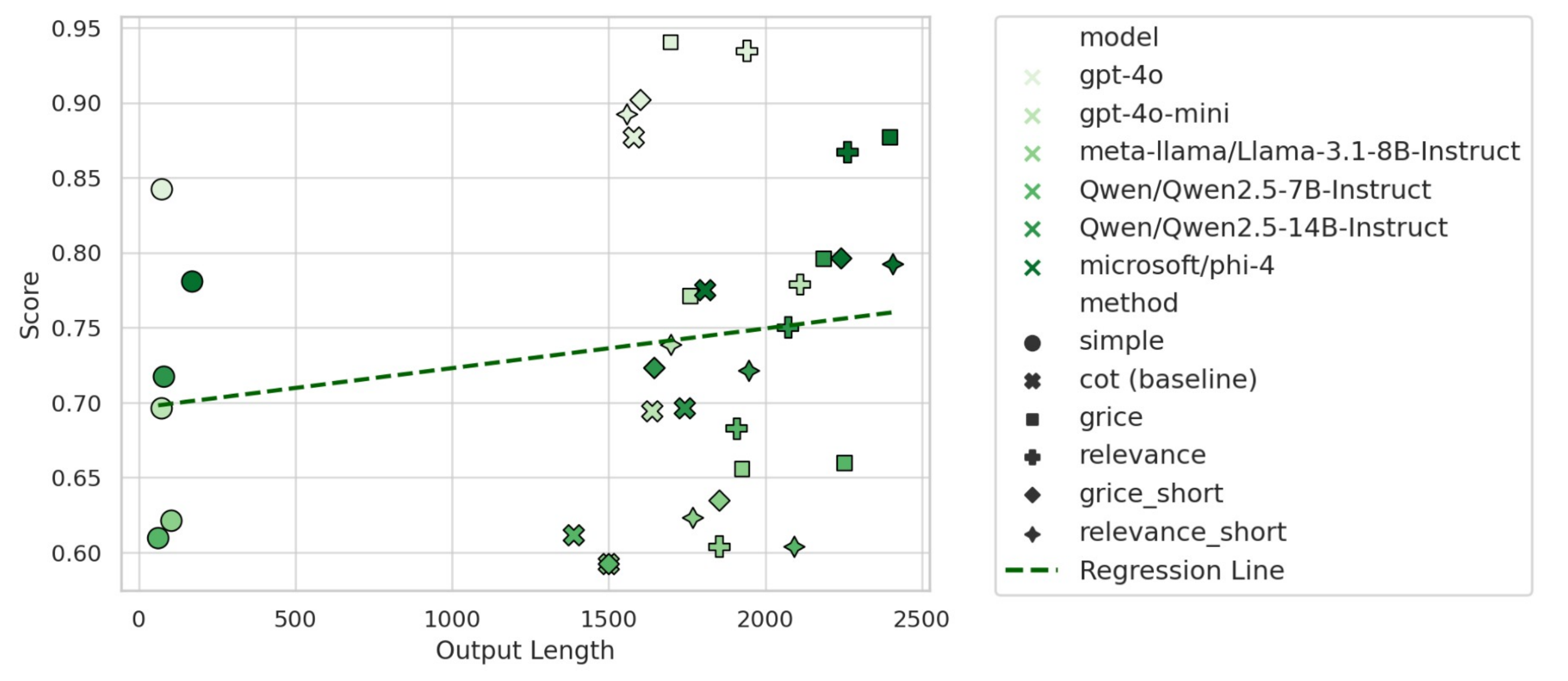}
        \caption{Correlation analysis between output length and accuracy.}
        \label{fig:correlation_analysis_output}
    \end{figure*}

The results of the analysis on the relationship between Accuracy scores and the length of Input to the model and the length of Output from the model are shown in Table \ref{fig:correlation_analysis_input} and \ref{fig:correlation_analysis_output}, respectively.

\section{Exact Results for Each Phenomena}
\label{appendix:exact_result_by_phenomona}
The exact experimental results aggregated for each phenomenon in PRAGMEGA are shown in Table \ref{tab:exact_result_by_phenometa_gpt4o}, \ref{tab:exact_result_by_phenometa_gpt4omini}, \ref{tab:exact_result_by_phenometa_llama3.1-8binstruct}, \ref{tab:exact_result_by_phenometa_qwen2.5-7binstruct}, \ref{tab:exact_result_by_phenometa_qwen2.5-14binstruct}, and \ref{tab:exact_result_by_phenometa_phi4}.

\begin{table*}[tbp]
    \centering
    \fontsize{9.1pt}{9.1pt}
    \caption{gpt-4o}
    \begin{tabularx}{\textwidth}{c|c|C|C|C|C|C}
        \hline
        & & \textbf{Deceits} & \textbf{Metaphor} & \textbf{Indirect Speech} & \textbf{Irony} & \textbf{Maxims} \\ \hline \hline

        \multirow{2}{*}{Baseline}
        & \cellcolor{black!5}\texttt{simple} & 0.97 & 0.88 & 0.93 & 0.69 & 0.76 \\ \cline{2-7}
        & \cellcolor{black!15}\texttt{cot} & 0.93 & 0.90 & 0.90 & 0.86 & 0.78 \\ \hline
        
        \multirow{2}{*}{Proposed} 
        & \cellcolor{yellow!5}\texttt{grice} & 0.97 & 0.96 & 0.95 & 0.92 & 0.89 \\ \cline{2-7}
        & \cellcolor{yellow!15}\texttt{relevance} & 0.94 & 0.98 & 0.97 & 0.92 & 0.85 \\ \hline

        \multirow{2}{*}{Short} 
        & \cellcolor{blue!5}\texttt{grice\_short} & 0.89 & 0.93 & 0.95 & 0.90 & 0.83 \\ \cline{2-7}
        & \cellcolor{blue!15}\texttt{relevance\_short} & 0.91 & 0.90 & 0.97 & 0.87 & 0.81 \\ \hline
    \end{tabularx}
    \label{tab:exact_result_by_phenometa_gpt4o}
\end{table*}

\begin{table*}[tbp]
    \centering
    \fontsize{9.1pt}{9.1pt}
    \caption{gpt-4o-mini}
    \begin{tabularx}{\linewidth}{c|c|C|C|C|C|C}
        \hline
        & & \textbf{Deceits} & \textbf{Metaphor} & \textbf{Indirect Speech} & \textbf{Irony} & \textbf{Maxims} \\ \hline \hline

        \multirow{2}{*}{Baseline}
        & \cellcolor{black!5}\texttt{simple} & 0.78 & 0.81 & 0.80 & 0.55 & 0.56 \\ \cline{2-7}
        & \cellcolor{black!15}\texttt{cot} & 0.74 & 0.79 & 0.74 & 0.66 & 0.53 \\ \hline
        
        \multirow{2}{*}{Proposed} 
        & \cellcolor{yellow!5}\texttt{grice} & 0.76 & 0.83 & 0.85 & 0.73 & 0.68 \\ \cline{2-7}
        & \cellcolor{yellow!15}\texttt{relevance} & 0.82 & 0.84 & 0.90 & 0.71 & 0.63 \\ \hline

        \multirow{2}{*}{Short} 
        & \cellcolor{blue!5}\texttt{grice\_short} & 0.69 & 0.83 & 0.83 & 0.64 & 0.64 \\ \cline{2-7}
        & \cellcolor{blue!15}\texttt{relevance\_short} & 0.79 & 0.82 & 0.87 & 0.70 & 0.50 \\ \hline
    \end{tabularx}
    \label{tab:exact_result_by_phenometa_gpt4omini}
\end{table*}

\begin{table*}[tbp]
    \centering
    \fontsize{9.1pt}{9.1pt}
    \caption{Llama-3.1-8B-Instruct}
    \begin{tabularx}{\linewidth}{c|c|C|C|C|C|C}
        \hline
        & & \textbf{Deceits} & \textbf{Metaphor} & \textbf{Indirect Speech} & \textbf{Irony} & \textbf{Maxims} \\ \hline \hline

        \multirow{2}{*}{Baseline}
        & \cellcolor{black!5}\texttt{simple} & 0.77 & 0.59 & 0.78 & 0.47 & 0.52 \\ \cline{2-7}
        & \cellcolor{black!15}\texttt{cot} & 0.70 & 0.62 & 0.74 & 0.51 & 0.40 \\ \hline
        
        \multirow{2}{*}{Proposed} 
        & \cellcolor{yellow!5}\texttt{grice} & 0.66 & 0.62 & 0.76 & 0.70 & 0.51 \\ \cline{2-7}
        & \cellcolor{yellow!15}\texttt{relevance} & 0.67 & 0.58 & 0.77 & 0.60 & 0.37 \\ \hline

        \multirow{2}{*}{Short} 
        & \cellcolor{blue!5}\texttt{grice\_short} & 0.68 & 0.57 & 0.76 & 0.65 & 0.49 \\ \cline{2-7}
        & \cellcolor{blue!15}\texttt{relevance\_short} & 0.71 & 0.60 & 0.77 & 0.58 & 0.45 \\ \hline
    \end{tabularx}
    \label{tab:exact_result_by_phenometa_llama3.1-8binstruct}
\end{table*}

\begin{table*}[tbp]
    \centering
    \fontsize{9.1pt}{9.1pt}
    \caption{Qwen2.5-7B-Instruct}
    \begin{tabularx}{\linewidth}{c|c|C|C|C|C|C}
        \hline
        & & \textbf{Deceits} & \textbf{Metaphor} & \textbf{Indirect Speech} & \textbf{Irony} & \textbf{Maxims} \\ \hline \hline

        \multirow{2}{*}{Baseline}
        & \cellcolor{black!5}\texttt{simple} & 0.82 & 0.55 & 0.78 & 0.40 & 0.52 \\ \cline{2-7}
        & \cellcolor{black!15}\texttt{cot} & 0.78 & 0.66 & 0.80 & 0.44 & 0.48 \\ \hline
        
        \multirow{2}{*}{Proposed} 
        & \cellcolor{yellow!5}\texttt{grice} & 0.72 & 0.64 & 0.90 & 0.55 & 0.49 \\ \cline{2-7}
        & \cellcolor{yellow!15}\texttt{relevance} & 0.77 & 0.65 & 0.86 & 0.51 & 0.44 \\ \hline

        \multirow{2}{*}{Short} 
        & \cellcolor{blue!5}\texttt{grice\_short} & 0.65 & 0.58 & 0.90 & 0.44 & 0.51 \\ \cline{2-7}
        & \cellcolor{blue!15}\texttt{relevance\_short} & 0.69 & 0.52 & 0.88 & 0.43 & 0.48 \\ \hline
    \end{tabularx}
    \label{tab:exact_result_by_phenometa_qwen2.5-7binstruct}
\end{table*}

\begin{table*}[tbp]
    \centering
    \fontsize{9.1pt}{9.1pt}
    \caption{Qwen2.5-14B-Instruct}
    \begin{tabularx}{\linewidth}{c|c|C|C|C|C|C}
        \hline
        & & \textbf{Deceits} & \textbf{Metaphor} & \textbf{Indirect Speech} & \textbf{Irony} & \textbf{Maxims} \\ \hline \hline

        \multirow{2}{*}{Baseline}
        & \cellcolor{black!5}\texttt{simple} & 0.90 & 0.69 & 0.76 & 0.64 & 0.61\\ \cline{2-7}
        & \cellcolor{black!15}\texttt{cot} & 0.85 & 0.70 & 0.78 & 0.56 & 0.53 \\ \hline
        
        \multirow{2}{*}{Proposed} 
        & \cellcolor{yellow!5}\texttt{grice} & 0.94 & 0.78 & 0.90 & 0.72 & 0.63 \\ \cline{2-7}
        & \cellcolor{yellow!15}\texttt{relevance} & 0.92 & 0.70 & 0.83 & 0.68 & 0.62 \\ \hline

        \multirow{2}{*}{Short} 
        & \cellcolor{blue!5}\texttt{grice\_short} & 0.84 & 0.72 & 0.87 & 0.63 & 0.56 \\ \cline{2-7}
        & \cellcolor{blue!15}\texttt{relevance\_short} & 0.81 & 0.73 & 0.88 & 0.57 & 0.58 \\ \hline
    \end{tabularx}
    \label{tab:exact_result_by_phenometa_qwen2.5-14binstruct}
\end{table*}

\begin{table*}[tbp]
    \centering
    \fontsize{9.1pt}{9.1pt}
    \caption{phi-4}
    \begin{tabularx}{\linewidth}{c|c|C|C|C|C|C}
        \hline
        & & \textbf{Deceits} & \textbf{Metaphor} & \textbf{Indirect Speech} & \textbf{Irony} & \textbf{Maxims} \\ \hline \hline

        \multirow{2}{*}{Baseline}
        & \cellcolor{black!5}\texttt{simple} & 0.88 & 0.79 & 0.84 & 0.74 & 0.65 \\ \cline{2-7}
        & \cellcolor{black!15}\texttt{cot} & 0.87 & 0.85 & 0.86 & 0.67 & 0.64 \\ \hline
        
        \multirow{2}{*}{Proposed} 
        & \cellcolor{yellow!5}\texttt{grice} & 0.93 & 0.86 & 0.94 & 0.90 & 0.73 \\ \cline{2-7}
        & \cellcolor{yellow!15}\texttt{relevance} & 0.90 & 0.87 & 0.90 & 0.84 & 0.67 \\ \hline

        \multirow{2}{*}{Short} 
        & \cellcolor{blue!5}\texttt{grice\_short} & 0.82 & 0.82 & 0.89 & 0.78 & 0.66 \\ \cline{2-7}
        & \cellcolor{blue!15}\texttt{relevance\_short} & 0.81 & 0.84 & 0.92 & 0.77 & 0.61 \\ \hline
    \end{tabularx}
    \label{tab:exact_result_by_phenometa_phi4}
\end{table*}

\section{Examples of Each Error Pattern}
\label{appendix:examples_of_each_error_pattern}
Table \ref{table:error_examples_in_appendix} shows specific examples of each error pattern by GPT-4o presented in Sec. \ref{subsubsection:error_analysis}.

    \begin{table*}[tbp]
        \centering
        \fontsize{9.1pt}{9.1pt}\selectfont
        \caption{
        Actual question examples for some error patterns.
        \color{blue}\textbf{Bold}\color{black} ~indicates the correct options.
        Due to space constraints, examples other than \textcircled{\scriptsize 1} and \textcircled{\scriptsize 3} are provided in Appendix \S\ref{appendix:examples_of_each_error_pattern}.
        }
        
        \begin{tabular}{m{1cm}|m{6cm}|m{8cm}}
        \hline
        \rowcolor{gray!10} 
        \textbf{Pattern} & \textbf{Questions} & \textbf{Options} \\ \hline \hline

        \textcircled{\scriptsize 1} & 
        Samantha is talking with her dad about her fiance. Samantha notes: ``John is an innocent person.'' Her dad replies: ``Undoubtedly, as innocent as a saint.'' Why has Samantha's dad responded like this? &1. Samantha's dad is impressed with John's innocence. \newline
        \textbf{\color{blue}2. Samantha's dad thinks that Samantha has an incorrect view of her fiance.\color{black}} \newline
        3. Samantha's dad thinks that Samantha's fiance is a saint. \newline
        4. Samantha's dad thinks that John is too religious. \\ \hline
        
        \textcircled{\scriptsize 2} & Lenny comes to the kitchen and asks his wife, Marcie: ``What will we have for breakfast?'' Marcie responds: ``A hard-boiled egg cooked in hot water and toast that is toasted evenly on both sides.'' Why has Marcie responded in such a way? & 1. Marcie is really good at cooking eggs and making toast.\newline 2. Marcie thinks that breakfast is the main meal of the day.\newline 3. Marcie wants Lenny to know how his breakfast was made.\newline \textbf{\color{blue}4. Marcie thinks that her husband's expectations about breakfast are too high.\color{black}} \\ \hline

        \textcircled{\scriptsize 3} & John is a teacher at an elementary school. When talking with the principal about a new student, who did poorly on her entrance examination, John said, ``This one is really sharp.'' What did John want to convey? & 1. The entrance exam is unfair. \newline 2. The pencils need to be sharpened. \newline 3. The student is smart. \newline \textbf{\color{blue}4. The student is not very clever.\color{black}} \\ \hline

        \textcircled{\scriptsize 4} & One day Jane comes home and is delighted to find her partner Anthony straightening up her apartment. Jane notices that Anthony threw out lots of things which were creating clutter, including an old photo that she had always kept on the coffee table. Anthony is worried that something is troubling Jane and asks if anything is wrong. Jane answers, ``Everything is fine, dear. You did a great job of cleaning the apartment.'' Why has Jane responded like this? & 1. She is happy that Anthony has cleaned the apartment and does not care about the picture that got thrown away.\newline 2. She wants to show that she is angry that Anthony has cleaned the apartment.\newline \textbf{\color{blue}3. She wants to show that she appreciates that Anthony has cleaned the apartment.\color{black}}\newline 4. She shows him how angry she is with him for throwing out things without her consent.\\ \hline

        \textcircled{\scriptsize 5} & Cindy wanted to paint a picture. She got her paints, paper and brushes ready. She has a meeting to go to in 10 minutes. Her dad said to her, ``I am not sure that now is the best time for painting.'' What might he be trying to convey? &\textbf{\color{blue}1. He does not want Cindy to start painting.\color{black}} \newline 2. He wants Cindy to create a sculpture.\newline 3. He wants Cindy to paint a picture for the meeting.\newline 4. He has some doubts whether Cindy should be painting.\newline  \\ \hline

        \end{tabular}
        \label{table:error_examples_in_appendix}
    \end{table*}

\end{document}